\documentclass[11pt,letterpaper]{article}

\usepackage{times}
\usepackage{url}
\usepackage[utf8x]{inputenc}
\usepackage[pdftex]{graphicx}
\usepackage{latexsym}
\usepackage{amsmath}

\usepackage{naaclhlt2013}

\title{Really? Well. Apparently Bootstrapping Improves the Performance of Sarcasm and Nastiness Classifiers for Online Dialogue}

\author{Stephanie Lukin\\
	    Natural Language and Dialogue Systems\\
	    University of California, Santa Cruz\\
	    1156 High Street, Santa Cruz, CA 95064\\
	    {\tt slukin@soe.ucsc.edu}
	  \And
	Marilyn Walker\\
  	Natural Language and Dialogue Systems\\
	University of California, Santa Cruz\\
	1156 High Street, Santa Cruz, CA 95064\\
  {\tt maw@soe.ucsc.edu}}

\date{}

\begin{document}
\maketitle
\begin{abstract}
More and more of the information  on the web is dialogic,
from Facebook newsfeeds, to forum conversations, to comment threads on
news articles.  In contrast to traditional, monologic Natural Language Processing 
resources such as
news, highly social dialogue is frequent in social media, making
it a challenging context for NLP.  This paper tests a bootstrapping
method, originally proposed in a monologic domain,
to train classifiers to identify two different types of subjective language in dialogue: 
sarcasm and nastiness. We explore two methods of developing
linguistic indicators to be used in a first level classifier
aimed at maximizing precision at the expense of recall.  The best
performing classifier for the first phase achieves 54\% precision and
38\% recall for sarcastic utterances. We then use general syntactic patterns
from previous work
to create more general sarcasm indicators, improving precision to 62\%
and recall to 52\%. To further test the generality of the method,
we then apply it  to bootstrapping a
classifier for nastiness dialogic acts. Our first phase, using
crowdsourced nasty indicators, achieves 58\% precision and 49\%
recall, which increases to 75\% precision and 62\% recall when we
bootstrap over the first level with generalized syntactic patterns.
\end{abstract}

\section{Introduction}
\label{intro-sec}

\begin{figure}[ht!b]
\begin{center}
\begin{scriptsize}
\begin{tabular}{|p{2.2in}|p{.2in}|p{.23in}|}
\hline  
Quote {\bf Q}, Response {\bf R} & {\bf Sarc} & {\bf Nasty} \\ \hline
{\bf Q1}: I jsut voted. sorry if some people actually have, you know, LIVES and don't sit around all day on debate forums to cater to some atheists posts that he thiks they should drop everything for. emoticon-rolleyes emoticon-rolleyes emoticon-rolleyes As to the rest of your post, well, from your attitude I can tell you are not Christian in the least. Therefore I am content in knowing where people that spew garbage like this will end up in the End. &  & \\
{\bf R1}: No, let me guess . . . er . . . McDonalds. No, Disneyland. Am I getting closer? & 1 & -3.6
\\ \hline  \hline 
{\bf Q2}: The key issue is that once children are born they are not physically dependent on a particular individual. &  &  \\
{\bf R2}  Really? Well, when I have a kid, I'll be sure to just leave it in the woods, since it can apparently care for itself. & 1 & -1 \\
\hline \hline 
{\bf Q3}: okay, well i think that you are just finding reasons to go against Him. I think that you had some bad experiances when you were younger or a while ago that made you turn on God. You are looking for reasons, not very good ones i might add, to convince people.....either way, God loves you. :) &  &  \\
{\bf R3}: Here come the Christians, thinking they can know everything by guessing, and commiting the genetic fallacy left and right. & 0.8 & -3.4 
\\ \hline  
\end{tabular}
\end{scriptsize}
\end{center}
\caption{\label{sample-quote-response} Sample Quote/Response Pairs
  from {\small \tt 4forums.com} with Mechanical Turk annotations for
  Sarcasm and Nasty/Nice. Highly negative values of Nasty/Nice
  indicate strong nastiness and sarcasm is indicated
by values near 1.}
\vspace{-.3in}
\end{figure}


More and more of the information on the web is dialogic, from Facebook
newsfeeds, to forum conversations, to comment threads on news
articles.  In contrast to traditional, monologic Natural Language Processing
resources such as news,
highly social dialogue is very frequent in social media, as
illustrated in the snippets in Fig.~\ref{sample-quote-response} from
the publicly available Internet Argument Corpus ({\bf IAC})
\cite{Walkeretal12c}.  Utterances are frequently sarcastic, e.g., {\it
  Really?  Well, when I have a kid, I’ll be sure to just leave it in
  the woods, since it can apparently care for itself} (R2 in
Fig.~\ref{sample-quote-response} as well as Q1 and R1), and are often
nasty, e.g.  {\it Here come the Christians, thinking they can know
  everything by guessing, and commiting the genetic fallacy left and
  right} (R3 in Fig.~\ref{sample-quote-response}).  Note also the
frequent use of dialogue specific discourse cues, e.g. the use of
{\it No} in R1, {\it Really? Well} in R2, and {\it okay, well} in Q3
in Fig.~\ref{sample-quote-response}~\cite{FoxTreeSchrock99,BryantFoxtree02,FoxTree10}.

The IAC comes with annotations of different types of social language 
categories including sarcastic vs not sarcastic, nasty vs nice, rational 
vs emotional and respectful vs insulting. Using a conservative
threshold of agreement amongst the annotators, an analysis of 
10,003 Quote/Response pairs (Q/R pairs) from the {\small \tt 4forums}
portion of IAC suggests that social subjective language is fairly
frequent: about 12\% of posts are sarcastic, 23\% are emotional, and
12\% are insulting or nasty.  We select sarcastic and nasty dialogic turns
to test our method on more than one type of subjective language and explore
issues of generalization; we do not claim any relationship between 
these types of social language in this work. 

Despite their frequency, expanding this corpus of sarcastic or 
nasty utterances at scale is expensive: human annotation of 
100\% of the corpus would be needed to identify 12\% more 
examples of sarcasm or nastiness. An explanation of how 
utterances are annotated in IAC is detailed in Sec.~\ref{prev-sec}.

Our aim in this paper is to explore whether it is possible to extend a
method for bootstrapping a classifier for monologic, subjective sentences proposed by Riloff \& Wiebe, henceforth R\&W
\cite{RiloffWiebe03,ThelenRiloff02}, to automatically find
sarcastic and nasty utterances in unannotated online dialogues. 
Sec.~\ref{method-sec} provides an overview of R\&W's
bootstrapping method. To apply bootstrapping, we: 
\begin{enumerate}
\item Explore two different methods for identifying cue words and phrases in 
two types of subjective language in dialogues: sarcasm and nasty  (Sec.~\ref{cue-sec}); 
\vspace{-.1in}
\item Use the learned indicators to train a sarcastic (nasty) dialogue act
  classifier that maximizes precision at the expense of recall (Sec.~\ref{hp-sec});
\vspace{-.1in}
\item Use the classified utterances to learn general
syntactic extraction patterns from the sarcastic (nasty) utterances (Sec.~\ref{pattern-sec});
\vspace{-.1in}
\item Bootstrap this process on unannotated text to learn
new extraction patterns to use for classification.
\vspace{-.1in}
\end{enumerate}

We show that the Extraction Pattern Learner improves
the precision of our sarcasm classifier by 17\% and the recall by
24\%, and improves the precision of the nastiness classifier
by 14\% and recall by 13\%.  We discuss previous work in 
Sec.~\ref{prev-sec} and compare to ours in Sec.~\ref{discuss-sec}
where we also summarize our results and discuss future work.

\section{Previous Work}
\label{prev-sec}

IAC provides
labels for sarcasm and nastiness that  were
collected with Mechanical Turk on Q/R pairs such as those in
Fig.~\ref{sample-quote-response}. Seven Turkers per Q/R pair
answered a {\bf binary} annotation question for
sarcasm {\it Is the respondent using sarcasm?} (0,1) and a {\bf scalar} annotation
question for nastiness {\it Is the respondent attempting to be nice or
  is their attitude fairly nasty?} (-5 nasty $\ldots$ 5 nice).  
We selected turns from IAC 
Table~\ref{tab-nums} with sarcasm averages
above 0.5, and nasty averages below -1 and nice above
1.  Fig.~\ref{sample-quote-response} included example nastiness and
sarcasm values.


Previous work on the automatic identification of sarcasm has focused on Twitter using the
{\tt \#sarcasm} \cite{gonzalezetal11} and {\tt \#irony} \cite{Reyesetal12} tags and 
a combined variety of tags and smileys \cite{davidovetal10}. Another popular
domain examines Amazon product reviews looking 
for irony \cite{ReyesRosso11}, sarcasm \cite{tsuretal10}, and a corpus
collection for sarcasm \cite{Filatova12}. \cite{Carvalhoetal09} looks for
irony in comments in online newpapers which can have a thread-like
structure. This primary focus on monologic 
venues suggests that sarcasm and irony can be detected with a relatively
high precision but have a different structure from dialogues 
\cite{FoxTreeSchrock99,BryantFoxtree02,FoxTree10}, posing the question, 
can we generalize from monologic to dialogic structures? 
Each of these works use methods including LIWC unigrams, affect, 
polarity, punctuation and more, and achieve on average a 
precision of 75\% or accuracy of between 45\% and 85\%.


Automatically identifying offensive utterances is also of interest. 
Previous work includes identifying flames in emails \cite{Spertus97}
and other messaging interfaces \cite{razavi2010offensive}, 
identifying insults in Twitter \cite{xiang2012detecting}, as well
as comments from new sites \cite{sood2011automatic}.
These approaches achieve an accuracy between 64\% and 83\%
using a variety of approaches. The accuracies for nasty utterances has 
a much smaller spread and 
higher average than sarcasm accuracies. This suggests that nasty
language may be easier to identify than sarcastic language.

\section{Method Overview}
\label{method-sec}

\begin{figure}[htb]
\centering
\includegraphics[width=3.25in]{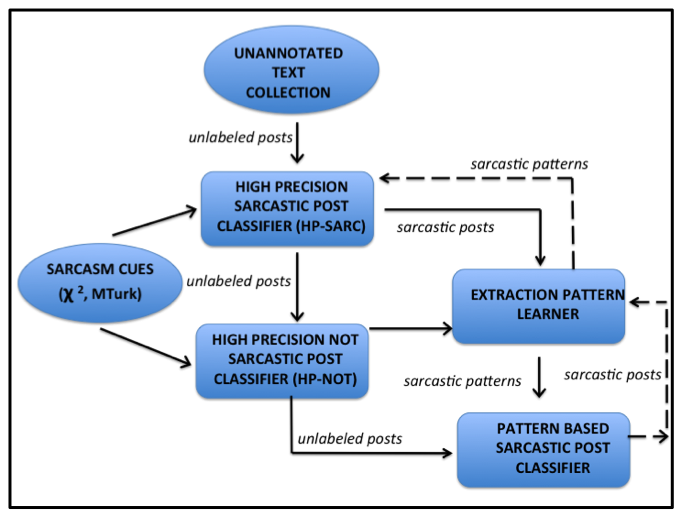}
\caption{\label{bootstrap-fig} Bootstrapping Flow for Classifying Subjective Dialogue Acts, shown for sarcasm, but identical for nastiness.}
\end{figure}

Our method for bootstrapping a classifier for sarcastic (nasty)
dialogue acts uses R\&W's model adapted to our data as 
illustrated for sarcasm in Fig.~\ref{bootstrap-fig}. 
The overall idea of the method is to find reliable cues and 
then generalize. 
The top of Fig.~\ref{bootstrap-fig} specifies the input to the method
as an unannotated corpus
of opinion dialogues, to illustrate the long term aim of building a
large corpus of the phenomenon of interest without human annotation.
Although the bootstrapping method assumes that the input is {\bf
  unannotated text}, we first need utterances that are already labeled for
sarcasm (nastiness) to train it.
Table~\ref{tab-nums} specifies how
we break down into datasets the annotations on the utterances in 
IAC for our various experiments.

\begin{table}[ht!]
\begin{small}
\begin{center}
\begin{tabular}{|p{0.75in}|c|c|c|}
\hline
SARCASM & \#sarc & \#notsarc & total    \\ \hline \hline
MT exp dev & 617  & NA & 617  \\ \hline
HP train  & 1407 & 1404 & 2811   \\ \hline
HP dev test   & 1614 & 1614 & 3228  \\ \hline
PE eval   & 1616 & 1616 & 3232  \\ \hline \hline
All &  5254 & 4635 & 9889   \\ \hline
\multicolumn{4}{c}{} \\ 
\multicolumn{4}{c}{} \\ \hline
NASTY &  \#nasty & \#nice & total   \\ \hline \hline
MT exp dev &  510 & NA & 510 \\ \hline
HP train  &  1147 & 1147 & 2294  \\ \hline
HP dev test &  691 & 691 & 1382 \\ \hline
PE eval &   691 & 691 & 1382 \\ \hline \hline
All   & 3039  & 2529 & 5568  \\ \hline
\end{tabular}
\caption{How utterances annotated for sarcasm (top) and nastiness (bottom) in IAC were used.
MT $=$ Mechanical Turk experimental development set. HP train $=$ utterances used to test whether combinations of cues could be used to develop a High
precision classifier. HP dev test $=$ ``Unannotated Text Collection'' in Fig.~\ref{bootstrap-fig}. 
PE eval $=$ utterances used to train the Pattern Classifier.
\label{tab-nums}} 
\end{center}
\end{small}
\end{table}

The left circle of Fig.~\ref{bootstrap-fig} reflects the assumption
that there are Sarcasm or Nasty Cues that can identify the
category of interest with high precision (R\&W call this the ``Known
Subjective Vocabulary''). The aim of first developing a high precision 
classifier, at the expense of recall, is to select utterances that are reliably 
of the category of interest from unannotated text. This is needed to 
ensure that the generalization step of  ``Extraction Pattern Learner"  does
not introduce too much noise.

R\&W did not need to develop a ``Known
Subjective Vocabulary'' because previous work provided one
\cite{Wilsonetal05,Wiebeetal99,Wiebeetal03}. Thus, our first question
with applying R\&W's method to our data was whether or not it is
possible to develop a reliable set of Sarcasm (Nastiness) Cues ({\bf O1} below). Two factors
suggest that it might not be. First, R\&W's
method assumes that the cues are in the utterance to be classified,
but it has been claimed that sarcasm (1) is context dependent, and (2)
requires world knowledge to recognize, at least in many cases.
Second, sarcasm is exhibited by a wide range of different forms
and with different dialogue strategies such as jocularity,
understatement and hyberbole
\cite{Gibbs00,Eisterholdetal06,BryantFoxtree02,Filatova12}.  In
Sec.~\ref{cue-sec} we devise and test two different methods for
acquiring a set of Sarcasm (Nastiness) Cues on particular development sets
of dialogue turns called the ``MT exp dev" in Table~\ref{tab-nums}.

The boxes labeled ``High Precision Sarcastic Post Classifier'' and
``High Precision Not Sarcastic Post Classifier'' in
Fig.~\ref{bootstrap-fig} involves using the Sarcasm
(Nastiness) Cues in simple combinations that 
maximize precision at the expense of recall. R\&W found
cue combinations that yielded a High Precision Classifier (HP Classifier)
with 90\% precision and 32\% recall on
their dataset. We discuss our test of these steps in
Sec.~\ref{hp-sec} on the ``HP train" development sets in
Table~\ref{tab-nums}
to estimate parameters for the High
Precision classifier, and then test the HP classifier with
these parameters on the test dataset labeled ``HP dev test" in 
Table~\ref{tab-nums}.

R\&W's Pattern Based classifier increased recall to 40\%
while losing very little precision. The open question with applying
R\&W's method to our data, was whether the cues that we discovered, by
whatever method, would work at high enough precision to support 
generalization ({\bf O2} below).  In
Sec.~\ref{pattern-sec} we describe how we use the ``PE eval" development set
(Table~\ref{tab-nums}) to estimate parameters for the Extraction
Pattern Learner, and then test the Pattern Based Sarcastic (Nasty) Post
classifier on the newly classified utterances
from the dataset labeled ``HP dev test" (Table~\ref{tab-nums}).
Our final open question was whether the extraction patterns from R\&W, which worked
well for news text, would work on social dialogue ({\bf O3} below). Thus 
our experiments address the following open questions as to whether R\&W's
bootstrapping method improves classifiers
for  sarcasm and nastiness in online dialogues:
\begin{itemize}
\item ({\bf O1}) Can we develop a ``known sarcastic (nasty) vocabulary''? The LH
circle of Fig.~\ref{bootstrap-fig} illustrates that we use
two different methods to identify {\bf Sarcasm Cues}. Because we have utterances labeled as sarcastic, we compare a statistical method that extracts important features automatically from utterances, with a method that has a human in the loop, asking annotators to select phrases that are good indicators of sarcasm (nastiness)
(Sec.~\ref{hp-sec});
\vspace{-.1in}
\item ({\bf O2}) If we can develop a reliable set of sarcasm (nastiness) cues,
is it then possible to develop an HP classifier? Will our precision be high enough? Is the fact that sarcasm is often context dependent an issue? 
(Sec.~\ref{hp-sec});
\vspace{-.1in}
\item ({\bf O3}) Will the extraction patterns used in R\&W's work 
allow us to generalize sarcasm cues from the HP Classifiers? 
Are R\&W's patterns 
general enough to work well for dialogue and social language? (Sec.~\ref{pattern-sec}).
\end{itemize}


\section{Sarcasm and Nastiness Cues}
\label{cue-sec}

\begin{table}[t] 
\begin{small}
\begin{center}
\begin{tabular}{|l|l|r|r|}
\hline
\multicolumn{4}{|c|}{\textbf{unigram}} \\ 
$\chi^2$ & MT & IA & FREQ \\ \hline \hline
right & ah & .95 & 2  \\ \hline
oh & relevant & .85 & 2 \\ \hline

we & amazing & .80 & 2 \\ \hline
same & haha & .75 & 2 \\ \hline
all  & yea & .73 & 3   \\ \hline
them & thanks & .68 & 6   \\ \hline
mean & oh & .56 & 56  \\ \hline \hline
\multicolumn{4}{|c|}{\textbf{bigram}} \\ 
 $\chi^2$ & MT   & IA & FREQ
\\ \hline \hline
the same & oh really & .83 & 2 \\ \hline
 mean like & oh yeah & .79 & 2 \\ \hline
trying to & so sure & .75 & 2 \\ \hline
that you & no way  & .72 & 3  \\ \hline
oh yeah & get real & .70 & 2  \\ \hline
I think & oh no & .66 & 4  \\ \hline 
we should & you claim  & .65 & 2  \\ \hline \hline
\multicolumn{4}{|c|}{\textbf{trigram}}\\ 
 $\chi^2$ & MT   & IA & FREQ
\\ \hline \hline
 you mean to & I get it & .97 & 3  \\ \hline 
mean to tell & I'm so sure & .65 & 2  \\ \hline 
have to worry & then of course & .65 & 2  \\ \hline 
sounds like a & are you saying  & .60 & 2  \\ \hline 
to deal with & well if you  & .55 & 2  \\ \hline 
I know I & go for it  & .52 & 2  \\ \hline 
you mean to & oh, sorry &  .50 & 2  \\ \hline 
\end{tabular}
\caption{Mechanical Turk (MT) and $\chi^2$ indicators for Sarcasm \label{fig:sarc_inds} }
\end{center}
\end{small}
\end{table}

\begin{table}[t] 
\begin{small}
\begin{center}
\begin{tabular}{|l|l|r|r|}
\hline

\multicolumn{4}{|c|}{\textbf{unigram}} \\ 
$\chi^2$ & MT & IA & FREQ \\ \hline \hline
like & idiot & .90 & 3 \\ \hline
them & unfounded & .85 & 2 \\ \hline
too & babbling & .80 & 2 \\ \hline
oh & lie & .72 & 11  \\ \hline
mean & selfish & .70 & 2 \\ \hline
just & nonsense & .69 & 9 \\ \hline
make & hurt & .67 & 3 \\ \hline \hline
\multicolumn{4}{|c|}{\textbf{bigram}} \\ 
 $\chi^2$ & MT   & IA & FREQ
\\ \hline \hline
 of the & don't expect & .95 & 2 \\ \hline
you mean & get your & .90 & 2 \\ \hline
yes, &   you're an & .85 & 2 \\ \hline 
 oh, & what's your & .77 & 4 \\ \hline
 you are & prove it &  .77 & 3 \\ \hline
 like a & get real & .75 & 2 \\ \hline
 I think & what else  & .70 & 2 \\ \hline  \hline
\multicolumn{4}{|c|}{\textbf{trigram}}\\ 
 $\chi^2$ & MT   & IA & FREQ
\\ \hline \hline
to tell me & get your sick   & .75 & 2  \\ \hline
would deny a & your ignorance is  &.70 & 2 \\ \hline
like that?  & make up your  & .70 & 2 \\ \hline
mean to tell & do you really  & .70 & 2 \\ \hline
sounds like a & do you actually &  .65 & 2 \\ \hline
you mean to & doesn't make it & .63 & 3 \\ \hline
to deal with & what's your point   & .60 & 2\\ \hline
\end{tabular}
\caption{Mechanical Turk (MT) and $\chi^2$ indicators for Nasty
\label{fig:nast_inds} }

\end{center}
\end{small}
\end{table}

Because there is no prior ``Known Sarcastic Vocabulary" we pilot
two different methods for discovering lexical cues to sarcasm and
nastiness, and experiment with combinations of cues that could yield
a high precision classifier \cite{Gianfortonietal11}. The first
method uses $\chi^2$ to measure whether a word or phrase is
statistically indicative of sarcasm (nastiness) in the development sets
labeled ``MT exp dev" (Table~\ref{tab-nums}). 
This method, a priori, seems reasonable because it is likely that if 
you have a large enough set of utterances labeled as sarcastic,
you could be able to automatically learn a set of reliable cues for sarcasm. 


The second method introduces a step of human annotation. 
We ask Turkers to identify
sarcastic (nasty) indicators in utterances (the open question {\bf O1}) from 
the development set ``MT exp dev" (Table~\ref{tab-nums}). 
Turkers were presented with utterances previously
labeled sarcastic or nasty in IAC by 7 different Turkers, 
and were told ``In a previous study, these responses were identified 
as being sarcastic by 3 out of 4 Turkers. For each quote/response pair, 
we will ask you to identify sarcastic or potentially sarcastic phrases 
in the response". The Turkers then selected words or phrases from the response 
they believed could lead someone to believing 
the utterance was sarcastic or nasty. These utterances were 
not used again in further experiments. This crowdsourcing method is 
similar to \cite{Filatova12}, but where their data is monologic, 
ours is dialogic.

\subsection{Results from Indicator Cues}

\begin{figure}[h]
\centering
\includegraphics[width=0.45\textwidth]{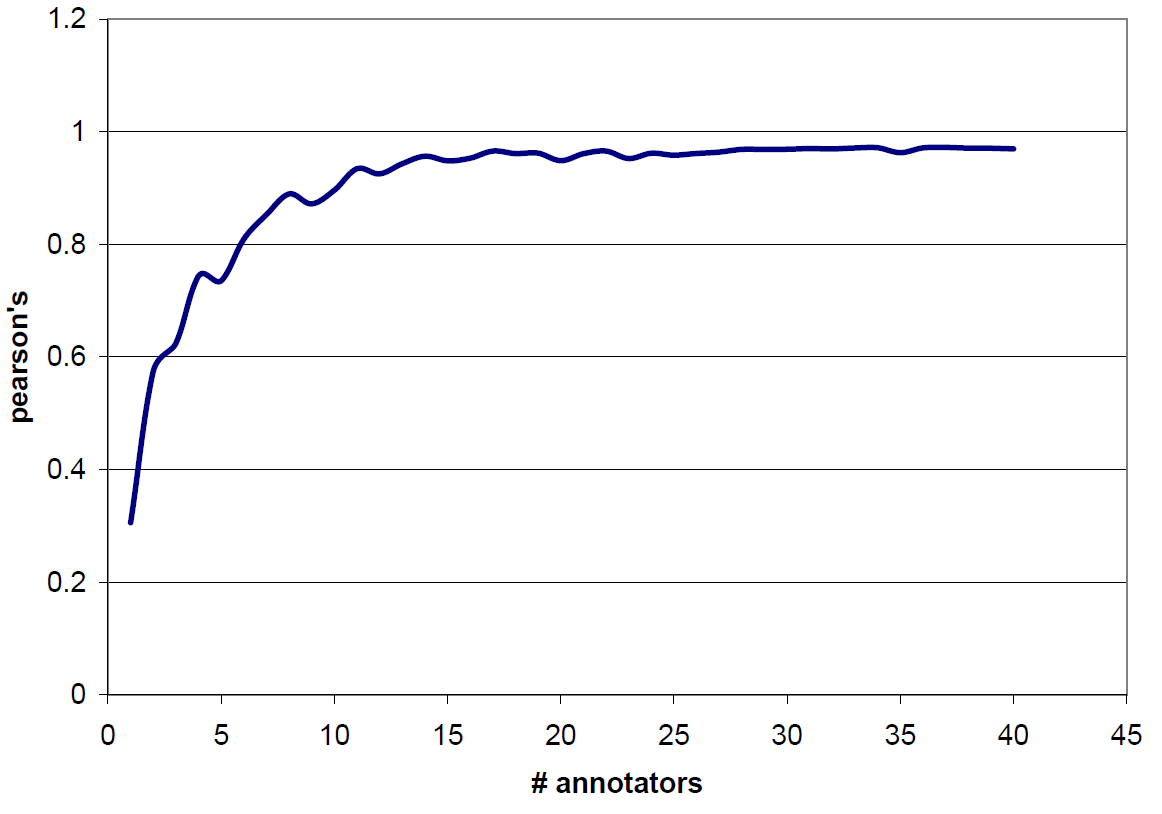}
\vspace{-.2in}
\caption{Interannotator Agreement for sarcasm trigrams}
\label{fig:tri40} \end{figure}

Sarcasm is known to be highly variable in form, and to depend, in some
cases, on context for its interpretation
\cite{SW81,Gibbs00,BryantFoxtree02}. We conducted an initial
pilot on 100 of the 617 sarcastic utterances in the development set
``MT exp dev" to see 
if this was necessarily the case in our dialogues. \cite{Snowetal08}
measures the quality of Mechanical Turk annotations on common 
NLP tasks by comparing them to 
a gold standard. Pearson's correlation coefficient shows that very few
Mechanical Turk annotators were required to beat the gold standard data, 
often less than 5. 
Because our sarcasm task does not have gold standard data, we ask 
100 annotators to participate in the pilot. Fig.~\ref{fig:tri40} plots the average 
interannotator agreement
(ITA) as a function of the number of annotators, computed using
Pearson correlation counts, for 40 annotators and for 
trigrams which require more data to converge. In all cases
(unigrams, bigrams, trigrams) ITA plateaus at around 20 annotators
and is about 90\% with 10 annotators, showing that the Mechanical Turk
tasks are well formed and there is high agreement. 
Thus we elicited only 10 annotations for the remainder of
the sarcastic and all the nasty utterances from the development 
set ``MT exp dev".

We begin to form our ``known sarcastic vocabulary" from these
indicators, (open question {\bf O1}). 
Each MT indicator has a {\bf FREQ} (frequency): the number
of times each indicator appears in the training set; and an {\bf IA} 
(interannotator agreement):
how many annotators agreed that each indicator was sarcastic or nasty.
Table~\ref{fig:sarc_inds} shows the best unigrams, bigrams, and
trigrams from the $\chi^2$ test and from the sarcasm Mechanical Turk
experiment and Table~\ref{fig:nast_inds} shows the results from the nasty 
experiment. We compare the MT
indicators to the $\chi^2$ indicators as part of investigating
open question {\bf O1}.  

As a pure statistical method, $\chi^2$ can pick out
things humans might not. For example, if it just happened that the word
`we' only occurs in sarcastic utterances in the development set, 
then $\chi^2$ will select it as a strong sarcastic word (row 3 of 
Table~\ref{fig:sarc_inds}). However, 
no human would recognize this word as corresponding to sarcasm. 
$\chi^2$ could easily be overtrained
if the ``MT exp dev" development set is not large enough to eliminate 
such general words from consideration, ``MT exp dev" only has 617 
sarcastic utterances and 510 nasty utterances (Table~\ref{tab-nums}).

Words that the annotators select as indicators (columns labeled MT in
Table~\ref{fig:sarc_inds} and Table~\ref{fig:nast_inds})
are much more easily identifiable although they do not
appear as often. For example, the {\bf IA} of 0.95 for
‘ah’ in Table~\ref{fig:sarc_inds} means that of all the annotators who saw ‘ah’ in the utterance
they annotated, 95\% selected it to be sarcastic. However the {\bf FREQ}
of 2 means that `ah’ only appeared in 2
utterances in the ``MT exp dev" development set. 

We test 
whether any of the methods for selecting indicators provide reliable
cues that generalize to a larger dataset in
Sec.~\ref{hp-sec}. The parameters that we estimate on the 
development sets are exactly how frequent (compared to a $\theta_1$) 
and how reliable (compared to a $\theta_2$) a cue
has to be to be useful in R\&W's bootstrapping method.

\section{High-Precision Classifiers}
\label{hp-sec}

R\&W use their ``known subjective vocabulary" to train a High
Precision classifier. R\&W's HP classifier searches for exact surface
matches of the subjective indicators and classifies utterances as
subjective if two subjective indicators are present.  We follow
similar guidelines to train HP Sarcasm and Nasty Classifiers. To test
open question {\bf O1}, we use a development set called
``HP train" (Table~\ref{tab-nums}) to test three methods for 
measuring the ``goodness'' of an
indicator that could serve as a high precision cue: (1) interannotator
agreement based on annotators consensus from Mechanical Turk, on
the assumption that the number of annotators that select a cue
indicates its strength and reliability ({\it IA features}); (2)
percent sarcastic (nasty) and frequency statistics in the HP train
dataset as R\&W do ({\it percent features}); and (3) the $\chi^2$ percent
sarcastic (nasty) and frequency statistics ({\it $\chi^2$ features}).

The {\it IA features} use the MT indicators and the {\bf IA} and {\bf FREQ} 
calculations introduced in Sec.~\ref{cue-sec}
(see Tables~\ref{fig:sarc_inds} and~\ref{fig:nast_inds}).
First, we select indicators 
such that $\theta_1 <= $ {\bf FREQ} where $\theta_1$ is a set of possible thresholds.
Then we introduce two new parameters $\alpha$ and $\beta$
to divide the indicators into three ``goodness'' groups that reflect 
interannotator agreement.
\begin{displaymath}
\begin{scriptsize}
   indicator strength = \left\{ 
  \begin{array}{l l}
    weak 		& \quad if \ 0 \le \text{{\bf IA}} < \alpha  \\
    medium	& \quad if \ \alpha \le \text{{\bf IA}} < \beta  \\
    strong		& \quad if \ \beta \le \text{{\bf IA}} < 1  \\
  \end{array} \right.
\end{scriptsize}
\end{displaymath}

For {\it IA features}, an utterance is classified as
sarcastic if it contains at least one {\it strong} or two {\it medium} indicators. Other conditions were piloted. We first
hypothesized that weak cues might be a way of classifying ``not
sarcastic'' utterances. But HP train showed that both sarcastic and 
not sarcastic utterances contain weak indicators yielding no
information gain. The same is true for Nasty’s counter-class Nice.  Thus 
we specify that counter-class utterances must have no 
{\it strong} indicators or at most one {\it medium} indicator. 
In contrast,  R\&W's counter-class classifier
looks for a maximum of one subjective indicator. 

The {\it percent features} also rely on the {\bf FREQ} of each MT
indicator, subject to a $\theta_1$ threshold, as well as the percentage of 
the time they occur in a sarcastic utterance ({\bf \%SARC}) or nasty utterance
({\bf \%NASTY}). We select indicators with various parameters for
$\theta_1$ and $\theta_2 \le $ {\bf \%SARC}. At least two indicators
must be present and above the thresholds to be classified and we
exhaust all combinations. Less than two indicators are needed to be
classified as the counter-class, as in R\&W.

Finally, the {\it $\chi^2$ features} use the same method as {\it percent 
features} only using the $\chi^2$ indicators instead of the MT indicators.

After determining which parameter settings performs the
best for each feature set, we ran the HP classifiers, 
using each feature set and the best
parameters, on the test set labeled ``HP dev test". 
The HP Classifiers classify the
utterances that it is confident on, and leave others unlabeled.

\subsection{Results from High Precision Classifiers}

The HP Sarcasm and Nasty Classifiers were trained on the 
three feature sets with the following parameters: {\it IA features} 
we exhaust all combinations of $\beta =$ [.70, .75, .80, .85, .90, .95, 1.00], 
$\alpha =$ [.35, .40, .45, .50, .55, .60, .65, .7],  and $\theta_1$ $=$ [2, 4, 6, 8, 10];
for the {\it percent features} and {\it $\chi^2$ features} we again
exhaust $\theta_1$ $=$ [2, 4, 6, 8, 10] and $\theta_2$ $=$ [.55, .60,
  .65, .70, .75, .80, .85, .90, .95, 1.00].

Tables~\ref{sarcasm-train_results} and ~\ref{nasty-train_results} show
a subset of the experiments with each feature set. We want to select
parameters that maximize precision without sacrificing too much
recall.  Of course, the parameters that yield the highest precision
also have the lowest recall, e.g.  Sarcasm {\it percent features},
parameters $\theta_1 = 4$ and $\theta_2 = 0.75$ achieve 92\% precision
but the recall is 1\% (Table~\ref{sarcasm-train_results}), and Nasty
{\it percent features} with parameters $\theta_1 = 8$ and $\theta_2 =
0.8$ achieves 98\% precision but a recall of 3\%
(Table~\ref{nasty-train_results}).  On the other end of the spectrum,
the parameters that achieve the highest recall yield a precision
equivalent to random chance.


\begin{table}[ht!]
\begin{scriptsize}
\begin{center}
\begin{tabular}{|p{0.40in}|r|c|c|r|}
\hline
SARC &  PARAMS  & P & R & N (tp) 
\\ \hline \hline
\% &  $\theta_1=$4, $\theta_2=$.55  & 62\% & 55\% &  768 \\ \hline  
 &  4, .6  & 72\% & 32\% &  458 \\ \hline 
 &  4, .65  & 84\% & 12\% &  170 \\ \hline 
 &  4, .75  & 92\% & 1\% &  23 \\ \hline \hline 

IA &  $\theta_1=$2, $\beta=$.90,  $\alpha=$.35  & 51\% & 73\% &   1,026 \\ \hline 
 &  2, .95,  .55  & 62\% & 13\% &  189 \\ \hline 
 &  2, .9, .55  & 54\% & 34\% &  472 \\ \hline 
 &  4, .75, .5  & 64\% & 7\% &  102 \\ \hline 
 &  4, .75, .6  & 78\% & 1\% &  22 \\ \hline \hline 

$\chi^2$ &  $\theta_1=$8, $\theta_2=$.55  & 59\% & 64\% &  893 \\ \hline 
 &  8, .6  & 67\% & 31\% &  434  \\ \hline  
 &  8, .65  & 70\% & 12\% &  170 \\ \hline 
 &  8, .75  & 93\% & 1\% &  14 \\ \hline 
\end{tabular}
\caption{Sarcasm Train results; P: precision, R: recall, tp: true positive classifications  
\label{sarcasm-train_results}}
\end{center}
\end{scriptsize}
\end{table}

\begin{table}[ht!]
\begin{scriptsize}
\begin{center}
\begin{tabular}{|p{0.40in}|r|c|c|r|}
\hline
NASTY &  PARAMS  & P & R & N (tp) 
\\ \hline \hline
\% &  $\theta_1=$2, $\theta_2=$.55  & 65\% & 69\% & 798\\ \hline 
 &  4, .65  & 80\% & 44\% & 509 \\ \hline 
 &  8, .75  & 95\% & 11\% & 125 \\ \hline  
 &  8, .8  & 98\% & 3\% & 45 \\ \hline \hline 

IA &  $\theta_1=$2, $\beta=$.95, $\alpha=$.35  & 50\% & 96\% &  1,126 \\ \hline 
 &  2, .95,  .45  & 60\% & 59\% & 693 \\ \hline 
 &  4, .75,   .45  & 60\% & 50\% & 580 \\ \hline 
 &  2, .7,   .55  & 73\% & 12\% & 149 \\ \hline 
 &  2, .9,   .65  & 85\% & 1\% & 17 \\ \hline \hline 

$\chi^2$ &   $\theta_1=$2, $\theta_2=$.55  & 73\% & 15\% & 187 \\ \hline 
 &  2, .65  & 78\% & 8\% & 104 \\ \hline   
 &  2, .7  & 86\% & 3\% & 32 \\ \hline     
\end{tabular}
\caption{Nasty Train results; P: precision, R: recall, tp: true positive classifications  
\label{nasty-train_results}}
\end{center}
\end{scriptsize}
\end{table}

Examining the parameter combinations
in Tables~\ref{sarcasm-train_results} and ~\ref{nasty-train_results} 
shows that {\it percent features} do better than {\it IA features} 
in all cases in terms of precision. Compare the block of results labeled \%
in Tables~~\ref{sarcasm-train_results} and \ref{nasty-train_results} 
with the IA and $\chi^2$ blocks for column P.
Nasty appears to be easier to identify than Sarcasm, especially using the {\it percent features}.
The performance of the {\it $\chi^2$ features} is comparable to that of
{\it percent features} for sarcasm, but lower than {\it percent features}
for Nasty. 

The best parameters selected from each feature set are shown in the
{\sc params} column of Table~\ref{hp_final_results}. With the
indicators learned from these parameters, we run the Classifiers on the test
set labeled ``HP
dev test" (Table~\ref{tab-nums}).  The performance on test set ``HP dev test"
(Table ~\ref{hp_final_results}) is worse than on the training set
(Tables ~\ref{sarcasm-train_results} and
~\ref{nasty-train_results}). However we conclude that {\bf both the
  \% and $\chi^2$ features} provide candidates for sarcasm (nastiness)
cues that are high enough precision (open question {\bf O2}) to be used in the
Extraction Pattern Learner (Sec.~\ref{pattern-sec}), even if Sarcasm
is more context dependent than Nastiness.

\begin{table}
\begin{scriptsize}
\begin{center}
\begin{tabular}{|p{0.45in}|r|c|c|c|c|}
\hline
& PARAMS & P & R & F    \\ \hline \hline
Sarc \% &  $\theta_1=$4, $\theta_2=$.55  & 54\% & 38\% & 0.46 \\ \hline
Sarc IA &  $\theta_1=$2,  $\beta=$.95,  $\alpha=$.55  & 56\% & 11\% & 0.34 \\ \hline
Sarc $\chi^2$ &  $\theta_1=$8, $\theta_2=$.60  & 60\% & 19\% & 0.40 \\ \hline \hline
Nasty \% &  $\theta_1=$2, $\theta_2=$.55  & 58\% & 49\% & 0.54 \\ \hline
Nasty IA &  $\theta_1=$2, $\beta=$.95, $\alpha=$.45  & 53\% & 35\% & 0.44 \\ \hline
Nasty $\chi^2$ &  $\theta_1=$2, $\theta_2=$.55  & 74\% & 14\% & 0.44 \\ \hline
\end{tabular}
\caption{HP Dev test results; {\sc PARAMS}: the best parameters for 
each feature set P: precision, R: recall, F: f-measure}
\label{hp_final_results} 
\end{center}
\end{scriptsize}
\end{table}

\section{Extraction Patterns}
\label{pattern-sec}


R\&W's Pattern Extractor searches for instances of the 13 templates in
the first column of Table~\ref{tab:templates} in utterances classified
by the HP Classifier. We reimplement this; an example of each
pattern as instantiated in test set ``HP dev test" for our data is shown in the
second column of Table~\ref{tab:templates}. The template {\tt \small
  <subj> active-verb <dobj>} matches utterances where a subject is
followed by an active verb and a direct object. However, these matches are not
limited to exact surface matches as the HP Classifiers required,
e.g. this pattern would match the phrase ``have a problem".
Table~\ref{extract-pattern-fig} in the Appendix provides example utterances
from IAC that match the instantiated template patterns.  
For example, the excerpt from the first row in Table~\ref{extract-pattern-fig}
``It is quite strange to encounter someone in this day and age who lacks any knowledge 
whatsoever of the mechanism of adaptation since it {\bf was explained} 150 years ago" 
matches the {\tt \small <subj> passive-verb} pattern. It appears 2 times ({\bf FREQ})
in the test set and is sarcastic both times ({\bf \%SARC} is 100\%). Row 11 in 
Table~\ref{extract-pattern-fig} shows an utterance matching the
{\tt \small active-verb prep <np>} pattern with the phrase 
``At the time of the Constitution there weren't exactly vast suburbs that 
could be prowled by thieves {\bf looking for} an open window". This phrase
appears 14 times ({\bf FREQ}) in the test set and is sarcastic ({\bf \%SARC}) 92\% of the
time it appears.

\begin{table}[ht!]
\begin{small}
\begin{center}
\begin{tabular}{|l|l|l|l|}
\hline
Synactic Form	&	Example Pattern	\\ \hline
$<$subj$>$ passive-verb	&	$<$subj$>$ was explained \\	
$<$subj$>$ active-verb	&	$<$subj$>$ appears	\\
$<$subj$>$ active-verb dobj	&	$<$subj$>$ have problem	\\
$<$subj$>$ verb infinitive	&	$<$subj$>$ have to do	\\
$<$subj$>$ aux noun	&	$<$subj$>$ is nothing	\\
active-verb $<$dobj$>$	&	gives $<$dobj$>$	\\
infinitive $<$dobj$>$	&	to force $<$dobj$>$	\\
verb infinitive $<$dobj$>$	&	want to take $<$dobj$>$	\\
noun aux $<$dobj$>$	&	fact is $<$dobj$>$		\\
noun prep $<$np$>$	&	argument against $<$np$>$\\	
active-verb prep $<$np$>$	&	looking for $<$np$>$	\\
passive-verb prep $<$np$>$	&	was put in $<$np$>$	\\
infinitive prep $<$np$>$	&	to go to $<$np$>$		\\ \hline
\end{tabular}
\caption{Syntactic Templates and Examples of Patterns that were Learned for Sarcasm. Table.~\ref{extract-pattern-fig} in the Appendix provides example posts that instantiate these patterns.}
\label{tab:templates} 
\end{center}
\end{small}
\end{table}

The Pattern Based Classifiers are trained on a development set labeled ``PE eval"
(Table~\ref{tab-nums}). Utterances from this development set are not used
again in any further experiments. Patterns are extracted from the dataset and we
again compute {\bf FREQ} and {\bf \%SARC} and {\bf \%NASTY} for each
pattern subject to $\theta_1 \le$ {\bf FREQ} and $\theta_2 \le$ {\bf
  \%SARC} or {\bf \% NASTY}.  Classifications are made if at least two
patterns are present and both are above the specified $\theta_1$ and
$\theta_2$, as in R\&W. Also following  R\&W, we do not learn
``not sarcastic" or ``nice" patterns.

To test the Pattern Based Classifiers, we use as input the
classifications made by the HP Classifiers. Using the predicted labels
from the classifiers as the true labels, the patterns from test set ``HP test dev"
are extracted and compared to those patterns found in development set ``PE eval". We have
two feature sets for both sarcasm and nastiness: one using the
predictions from the MT indicators in the HP classifier ({\it percent
  features}) and another using those instances from the {\it $\chi^2$
  features}.


\subsection{Results from Pattern Classifier}

\begin{figure}[t] 
\centering
\includegraphics[width=.5\textwidth]{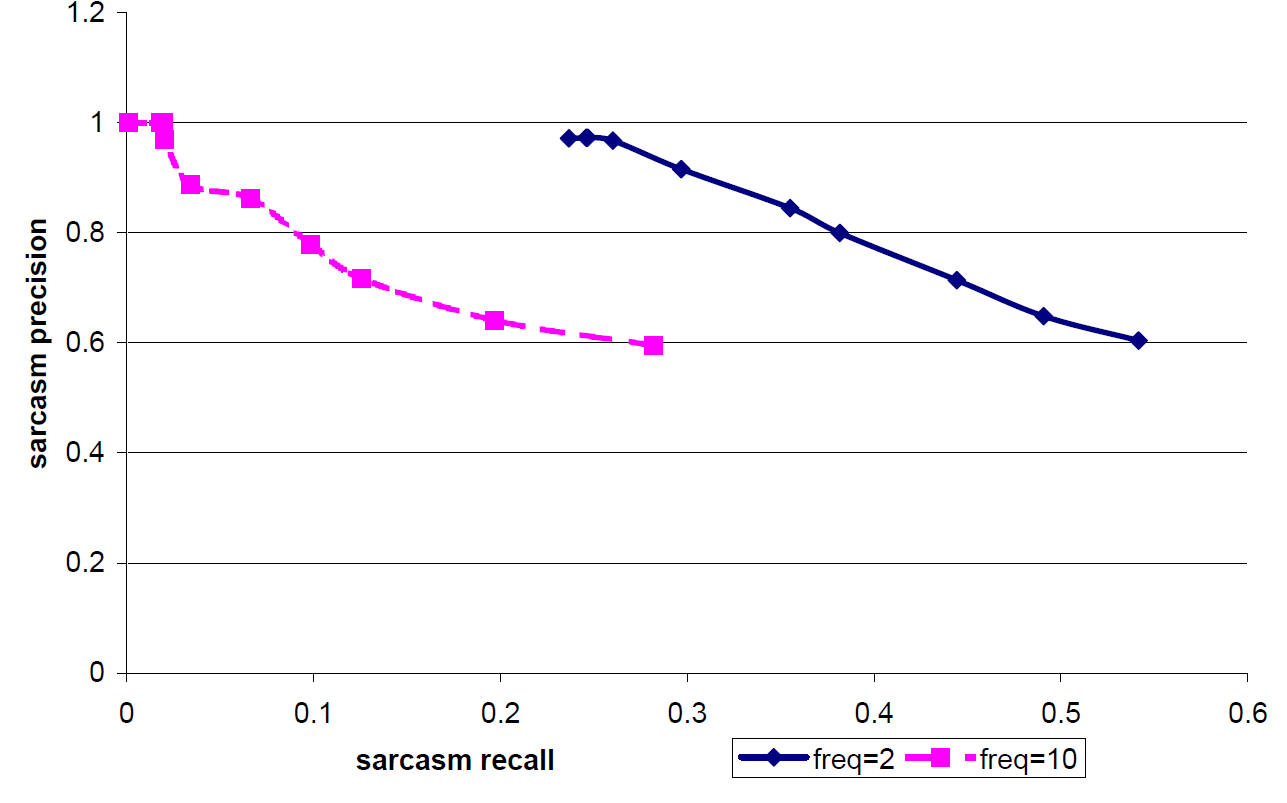}
\caption{Recall vs. Precision for Sarcasm PE eval \label{param-fig}}
\end{figure}

The Pattern Classifiers classify an utterance as Sarcastic (Nasty) if
at least two patterns are present and above the thresholds $\theta_1$
and $\theta_2$, exhausting all combinations of $\theta_1$ $=$ [2, 4,
  6, 8, 10] and $\theta_2$ $=$ [.55, .60, .65, .70, .75, .80, .85,
  .90, .95, 1.00]. The counter-classes are predicted when the
utterance contains less than two patterns.  The exhaustive classifications 
are first made using the utterances in the development set labeled ``PE eval".
Fig.~\ref{param-fig} shows the precision and recall trade-off
for $\theta_1 =$ [2, 10] and all $\theta_2$ values on sarcasm development set``PE
eval".  As recall increases, precision drops. By including patterns
that only appear 2 times, we get better recall. Limiting $\theta_1$ to
10 yields fewer patterns and lower recall.

\pagebreak
Table~\ref{pc_eval} shows the results for various parameters.
The PE dev dataset learned a total of 1,896 sarcastic 
extraction patterns above a minimum threshold of $\theta_1 < 2$ and
$\theta_2 < 0.55$, and similarly 847 nasty extraction patterns. 
Training on development set ``PE dev" yields high precision and
good recall. To select the best parameters, we again look for a balance between 
precision and recall. Both Classifiers have very high precision. In the end,
we select parameters that have a better recall than the best parameter
from the HP Classifiers which is $recall = 38\%$ for sarcasm and $recall = 49\%$
for nastiness. The best parameters and their test results are shown in 
 Table~\ref{pc_final_resuts}.

\begin{table}[ht!]
\begin{scriptsize}
\begin{center}
\begin{tabular}{|p{0.40in}|r|c|c|c|c|}
\hline
 &  PARAMS  & P & R & F & N (tp) 
\\ \hline \hline
SARC &  $\theta_1=$2, $\theta_2=$.60  & 65\% & 49\% & 0.57 & 792\\ \hline 
 &  2, .65  & 71\% & 44\% &  0.58 & 717\\ \hline 
 &  2, .70  & 80\% & 38\% & 0.59 & 616\\ \hline  
 &  2, 1.0  & 97\% & 24\% & 0.60 & 382 \\ \hline  
\multicolumn{6}{c}{} \\  \hline
NASTY &  $\theta_1=$2, $\theta_2=$.65  & 71\% & 49\% & 0.60 & 335\\ \hline 
 &  2, .75  & 83\% & 42\% & 0.62 & 289 \\ \hline  
 &  2, .90  & 96\% & 30\% & 0.63 & 209 \\ \hline 

\end{tabular}
\caption{Pattern Classification Training; P: precision, R: recall, F: F-measure, tp: true positive classifications  
\label{pc_eval}}
\end{center}
\end{scriptsize}
\end{table}

The Pattern Classifiers are tested on ``HP dev test" with the labels
predicted by our HP Classifiers, thus we have two different sets of
classifications for both Sarcasm and Nastiness: {\it percent features}
and {\it $\chi^2$ features}. Overall, the Pattern Classification
performs better on Nasty than Sarcasm.  Also, the {\it percent
  features} yield better results than $\chi^2$ features, possibly
because the precision for $\chi^2$ is high from the HP Classifiers, but the recall is
very low.  We believe that $\chi^2$ selects statistically predictive
indicators that are tuned to the dataset, rather than general.  Having
{\bf a human in the loop guarantees more general features} from a smaller dataset.
Whether this remains true on the size as the dataset increases to 1000 or more
is unknown. We
conclude that R\&W's patterns generalize well on our Sarcasm and Nasty
datasets (open question {\bf O3}), but suspect that there may be 
better syntactic patterns for
bootstrapping sarcasm and nastiness, e.g. involving cue words or
semantic categories of words rather than syntactic categories, as we
discuss in Sec.~\ref{discuss-sec}. 

\begin{table}[ht!]
\begin{small}
\begin{center}
\begin{tabular}{|p{0.5in}|r|c|c|c|c|}
\hline
& PARAMS & P & R & F    \\ \hline \hline
Sarc \% & $\theta_1=$2, $\theta_2=$.70 &   62\% & 52\% & 0.57 \\ \hline
Sarc $\chi^2$ & $\theta_1=$2, $\theta_2=$.70  &   31\% & 58\% & 0.45 \\ \hline \hline
Nasty \%  & $\theta_1=$2, $\theta_2=$.65 &  75\% & 62\% & 0.69 \\ \hline
Nasty $\chi^2$ & $\theta_1=$2, $\theta_2=$.65  & 30\% & 70\% & 0.50 \\ \hline
\end{tabular}
\caption{The results for Pattern Classification on HP dev test dataset \label{pc_final_resuts};
 PARAMS: the best parameters for 
each feature set P: precision, R: recall, F: f-measure}
\end{center}
\end{small}
\end{table}

This process can be repeated by taking the newly classified utterances from
the Pattern Based Classifiers, then applying the Pattern Extractor 
to learn new patterns from the
newly classified data. This can be repeated for multiple iterations.
We leave this  for future work.



\section{Discussion and Future Work}
\label{discuss-sec}

In this work, we apply a bootstrapping method to train classifiers to identify
particular types of subjective utterances in online dialogues.  
First we create a suite of linguistic indicators for sarcasm 
and nastiness using crowdsourcing techniques. Our crowdsourcing method is 
similar to \cite{Filatova12}. From these new linguistic indicators we construct a classifier
following previous work on bootstrapping subjectivity classifiers 
\cite{RiloffWiebe03,ThelenRiloff02}. We compare the performance of 
the High Precision Classifier that was trained based on statistical measures
against one that keeps human annotators in the loop, and find that Classifiers using statistically selected 
indicators appear to be overtrained on the development set because they 
do not generalize well. This first phase achieves 
54\% precision and 38\% recall for sarcastic utterances using the human 
selected indicators. If we bootstrap by using syntactic patterns to create more
general sarcasm indicators from the utterances identified as sarcastic
in the first phase, we achieve a higher precision of 62\% and recall
of 52\%. 

We apply the same method to bootstrapping a classifier for nastiness
dialogic acts. Our first phase, using crowdsourced nasty indicators,
achieves 58\% precision and 49\% recall, which increases to 75\%
precision and 62\% recall when we bootstrap with syntactic patterns,
possibly suggesting that nastiness (insults) are less nuanced and
easier to detect than sarcasm. 

Previous work claims that recognition
of sarcasm (1) depends on knowledge of the speaker, (2) world
knowledge, or (3) use of context
\cite{Gibbs00,Eisterholdetal06,BryantFoxtree02,Carvalhoetal09}.
While we also believe that certain types of subjective language cannot
be determined from cue words alone, our Pattern Based Classifiers,
based on syntactic patterns, still achieves high precision and recall.
In comparison to previous monologic works whose sarcasm precision
is about 75\%, ours is not quite as good with 62\%. While the nasty works
do not report precision, we believe that they are comparable to the 
64\% - 83\% accuracy with our precision of 75\%.

Open question {\bf O3} was whether R\&W's patterns are fine tuned
to subjective utterances in news. However R\&W's patterns improve both
precision and recall of our Sarcastic and Nasty classifiers. In future
work however, we would like to test whether semantic categories of
words rather than syntactic categories would perform even better for
our problem, e.g. Linguistic Inquiry and Word Count categories.  
Looking again at row 1 in 
Table~\ref{extract-pattern-fig}, ``It is quite strange to encounter someone in 
this day and age who lacks any knowledge whatsoever of the mechanism of 
adaptation since it was explained 150 years ago", the word `quite' matches the 
`cogmech' and `tentative' categories, which might be interesting to generalize to sarcasm. 
In row 11 ``At the time of the Constitution there weren't exactly vast suburbs that 
could be prowled by thieves looking for an open window", the phrase
``weren't exactly" could also match the LIWC categories `cogmech' and `certain' or, 
more specifically, certainty negated.

We also plan to extend this work to other categories of subjective 
dialogue acts, e.g. emotional and respectful as mentioned in the 
Introduction, and to expand our corpus
of subjective dialogue acts. We will experiment with performing more
than one iteration of the bootstrapping process (R\&W complete two
iterations) as well as create a Hybrid Classifier combining the
subjective cues and patterns into a single Classifier that itself can
be bootstrapped.

Finally, we would like to extend our method to different dialogue domains
to see if the classifiers trained on our sarcastic and nasty indicators would 
achieve similar results or if different social media sites have their own style
of displaying sarcasm or nastiness not comparable to those in forum debates. 

\bibliographystyle{naaclhlt2013}
\bibliography{nl}

\section{Appendix A. Instances of  Learned Patterns}~~\begin{table*} [h]
\begin{small}
\begin{center}
\begin{tabular}{|p{1.70in}|c|c||p{4.0in}|}
\hline
Pattern Instance & FREQ & \%SARC & Example Utterance \\ \hline \hline
{\tt \footnotesize    <subj> was explained} & 2 & 100\% &  Well, I incorrectly assumed that anyone attempting to enter the discussion would at least have a grasp of the most fundamental principles. It is quite strange to encounter someone in this day and age who lacks any knowledge whatsoever of the mechanism of adaptation since it {\bf was explained} 150 years ago. \\ \hline
{\tt \footnotesize    <subj> appears} & 1 & 94\% & It {\bf appears} this thread has been attacked by the ``line item '' poster. \\ \hline
{\tt \footnotesize    <subj> have problem} & 4 & 50\% &  I see your point, Iangb but I'm not about to be leaving before you've had a chance to respond. I won't be    "leaving   " at all. You challenged me to produce an argument, so I'm going to produce my argument. I will then summarize the argument, and you can respond to it and we can then discuss / debate those specifics that you {\bf have a problem} with. \\ \hline
{\tt \footnotesize    <subj> have to do} & 15 & 86\% &  How does purchasing a house {\bf have to do} with abortion? Ok, so what if the kid wants to have the baby and the adults want to get rid of it? What if the adults want her to have the baby and the kid wants to get rid of it? You would force the kid to have a child (that doesn't seem responsible at all), or you would force the kid to abort her child (thereby taking away her son or daughter). Both of those decisions don't sound very consitent or responsible. The decision is best left up to the person that is pregnant, regardless of their age. \\ \hline
{\tt \footnotesize    <subj> is nothing} & 10 & 90\% &  Even though there {\bf is nothing} but ad hoc answers to the questions, creationists touted the book as \   "proof\   " that Noah\'s ark was possible. They never seem to notice that no one has ever tried to build and float an ark. They prefer to put the money into creation museums and amusement parks.  \\ \hline
{\tt \footnotesize    gives <dobj>} & 25 & 88\% & Just knowing that there are many Senators and Congressmen who would like to abolish gun rights {\bf gives} credence to the fact that government could actually try to limit or ban the 2nd Amendment in the future. \\ \hline
{\tt \footnotesize    to force <dobj>} & 9 & 89\% &  And I just say that it would be unjust and unfair of you {\bf to force} metaphysical belief systems of your own which constitute religious belief upon your follows who may believe otherwise than you. Get pregnant and treat your fetus as a full person if you wish, nobody will force you to abort it. Let others follow their own beliefs differing or the same. Otherwise you attempt to obtain justice by doing injustice  \\ \hline
{\tt \footnotesize    want to take <dobj>} & 5 & 80\% & How far do you {\bf want to take} the preemptive strike thing? Should we make it illegal for people to gather in public in groups of two or larger because anything else might be considered a violent mob assembly for the basis of creating terror and chaos? \\ \hline
{\tt \footnotesize    fact is <dobj>} & 6 & 83\% &  No, the {\bf fact is} PP was founded by an avowed racist and staunch supporter of Eugenics.  \\ \hline
{\tt \footnotesize    argument against <np>} & 4 & 75\% & Perhaps I am too attached to this particular debate that you are having but if you actually have a sensible {\bf argument against} gay marriage then please give it your best shot here. I look forward to reading your comments. \\ \hline
{\tt \footnotesize    looking for <np>} & 14 &  92\% & At the time of the Constitution there weren't exactly vast suburbs that could be prowled by thieves {\bf looking for} an open window. \\ \hline
{\tt \footnotesize    was put in <np>} & 3 & 66\% &  You got it wrong Daewoo. The ban {\bf was put in} place by the 1986 Firearm Owners Protection Act, designed to correct the erronius Gun Control Act of 1968. The machinegun ban provision was slipped in at the last minute, during a time when those that would oppose it weren\'t there to debate it. \\ \hline
{\tt \footnotesize    to go to <np>} & 8 & 63\% &  Yes that would solve the problem wouldn't it,worked the first time around,I say that because we (U.S.)are compared to the wild west. But be they whites,Blacks,Reds,or pi** purple shoot a few that try to detain or threaten you, yeah I think they will back off unless they are prepared {\bf to go to} war.  \\ \hline
\end{tabular}
\caption{Sarcastic patterns and example instances \label{extract-pattern-fig}}
\end{center}
\end{small}
\end{table*}

\end{document}